\documentclass[letterpaper, 10 pt, journal, twoside]{IEEEtran}

\IEEEoverridecommandlockouts                              

\makeatletter
\let\NAT@parse\undefined
\makeatother
\usepackage[sort&compress, numbers]{natbib}

\usepackage[pdftex]{graphicx}
\usepackage{booktabs}
\usepackage{threeparttable}
\usepackage{relsize}
\usepackage{amsmath}
\usepackage{amssymb}  
\usepackage[caption=false,font=footnotesize]{subfig}
\usepackage{capt-of}
\usepackage{multirow}
\usepackage{array,booktabs}
\usepackage{diagbox}
\usepackage{balance}
\usepackage[ruled]{algorithm}
\usepackage{algpseudocode}
\usepackage{float}
\usepackage{tabularx}
\usepackage{bm}
\usepackage{pifont}
\newcommand{\cmark}{\ding{51}}
\newcommand{\xmark}{\ding{55}}

 \usepackage[final]{hyperref}
 \hypersetup{
     colorlinks=true,
     linkcolor=blue,
     filecolor=magenta,
     urlcolor=blue,
     citecolor=black
 }
\usepackage[table]{xcolor}
\usepackage{./rpm_packages/rpm_SIunits}
\usepackage{./rpm_packages/rpm_acronyms}
\usepackage{./rpm_packages/rpm_math}
\usepackage{./rpm_packages/rpm_misc}

\usepackage{soul,color}
\usepackage{lipsum}

\usepackage{xcolor}

\usepackage{tikz} 

\title{RaDiVe: Robust 4D Radar Odometry with Distance-Bounded NDT and Velocity-Discrepancy Point Uncertainty}

\author{Sangwoo Jung${}^{1}$, Dongjae Lee${}^{1}$, Chiyun Noh${}^{1}$, and Ayoung Kim${}^{1*}$
\thanks{Manuscript received: July 30, 2026; Revised: October 21, 2026; Accepted: November 12, 2026. This paper was recommended for publication by Editor Teresa A. Vidal-Calleja upon evaluation of the Associate Editor and Reviewers’ comments. \textit{(Corresponding author: Ayoung Kim)}}
\thanks{This work was supported in part by the National Research Foundation of Korea (NRF) grant funded by the Korean government (MSIT) in 2024 (RS-2024-00461409, Development of Harmonic-space Operation using Seamless Information Sharing among Large-scale Heterogeneous Unmanned Vehicles), and in part by the Robotics and AI (RAI) Institute.}
\thanks{$^{1}$S. Jung, D. Lee, C. Noh, and A. Kim are with the Department of Mechanical Engineering, SNU, Seoul, S. Korea {\tt\small [dan0130, pur22, gch06208, ayoungk]@snu.ac.kr}}
\thanks{The source code is available at https://github.com/SangwooJung98/RaDiVe.}
\thanks{Digital Object Identifier (DOI): see top of this page.}
}

\begin{document}

\maketitle

\begin{abstract}

Recent advances in 4D radar enable robust perception in adverse weather; however, the inherent sparsity, noise, and limited positional precision of radar point clouds pose significant challenges for registration-based odometry. In this letter, we propose RaDiVe, a 4D radar odometry framework that improves the accuracy and robustness of radar point cloud registration. We introduce a distance-bounded Normal Distributions Transform, which improves optimization stability and computational efficiency by restricting the correspondence search to near-distance voxel pairs. To mitigate measurement ambiguity, we propose a velocity-discrepancy point uncertainty model that weights each input 4D radar point according to the discrepancy between its measured Doppler radial velocity and the radial velocity predicted from the estimated ego-velocity. We also incorporate Signed Distance Function-based surface point extraction via implicit neural mapping to construct a geometrically consistent, noise-filtered local submap. Evaluations across multiple public datasets demonstrate that RaDiVe outperforms existing 4D radar odometry baselines by 44.4\% in translational and 21.3\% in rotational Absolute Trajectory Error on average, while maintaining real-time performance. 
\end{abstract}




\section{Introduction}
\label{sec:intro}

\IEEEPARstart{R}{adar} serves as a key sensor for robust odometry in challenging environments due to its all-weather perception capabilities \cite{kellner2013instantaneous, doer2020ekf, park20213d, michalczyk2022tightly, zhuang20234d, jung2024co, noh2026garlileo}. As a range-sensing modality similar to \ac{LiDAR}, 4D radar-motivated approaches adopt \ac{LiDAR}-style scan-to-submap registration to incorporate geometric information into odometry. Nevertheless, the inherent sparsity, limited precision, and high noise characteristic of 4D radar point clouds degrade odometry accuracy, even with recent efforts to increase point throughput.

Despite these data quality challenges, \ac{ICP} is widely explored for existing 4D radar odometry approaches \cite{zhang20234dradarslam, herraez2024radar, kim2024radar4motion, herraez2025ground, yang2025ground, xu2025incorporating}. However, since \ac{ICP} inherently relies on point-level correspondences and assumes dense, high-precision measurements, its mismatch with the sparse, imprecise, and noisy nature of 4D radar point clouds often results in incorrect point correspondences. As an alternative to \ac{ICP}, \ac{NDT} registration has attracted attention for robustness on noisy data due to its voxel-based representation and smooth non-linear likelihood surfaces \cite{kung2021normal,zhang2023scan,hilger2024randt}. Previous \ac{NDT}-based approaches have proven effective on \ac{BEV} radar images; however, extending them to 4D radar point clouds remains challenging because fewer points per voxel yield unreliable voxel statistics and incorrect voxel correspondences. 

\begin{figure}[!t]
    \centering
    \includegraphics[width=0.9\columnwidth]{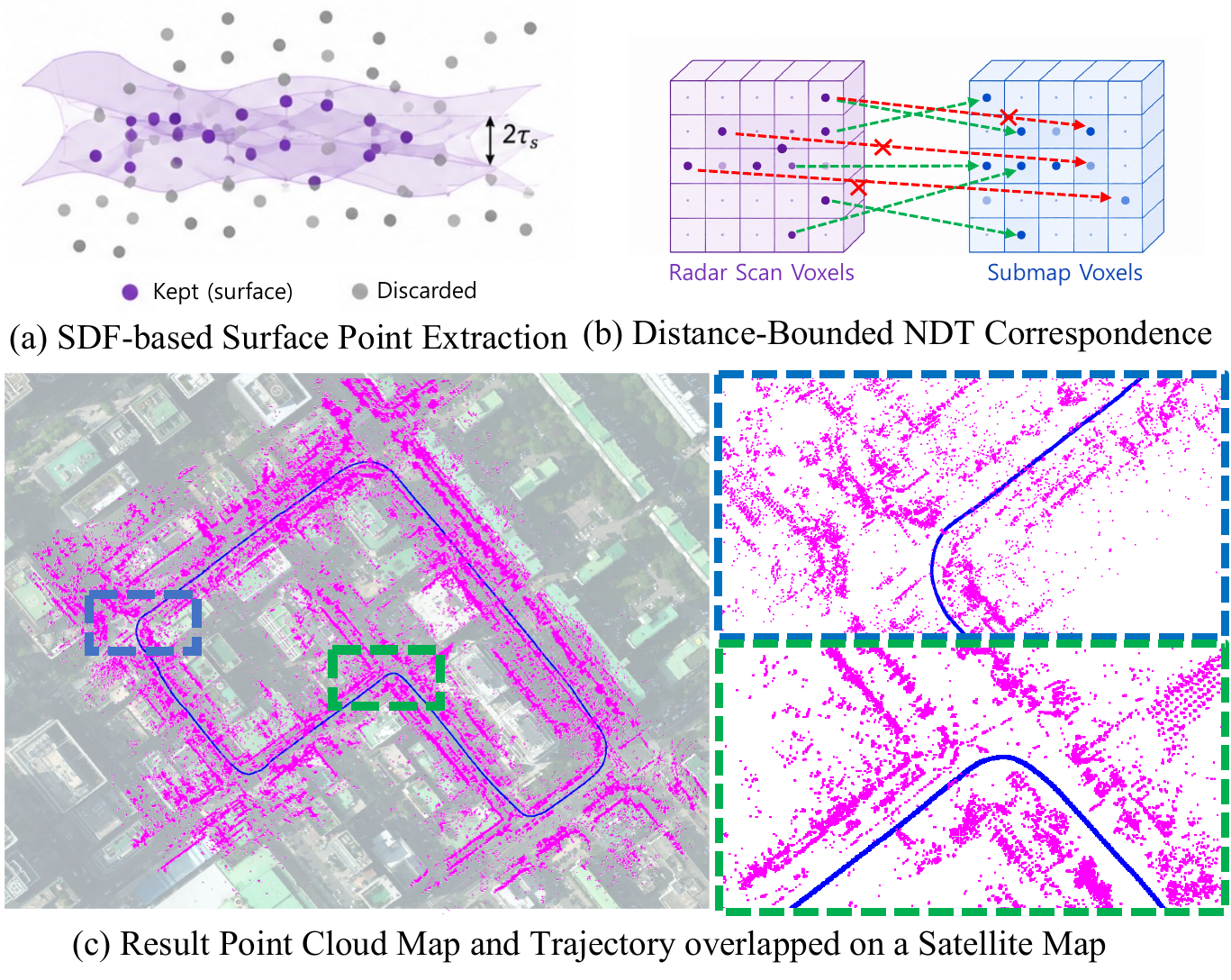}
    \vspace{-4mm}
    \caption{(a) Neural points closer than threshold $\tau_s$ to the surface are kept, while others are discarded. (b) \ac{NDT} correspondences are restricted to nearby pairs to enhance optimization stability and efficiency. (c) Qualitative results of RaDiVe on the \texttt{Street Day 1} sequence of \texttt{HeRCULES}. (Left) Neural point cloud map (purple) projected into a Google Maps satellite view. (Right) Zoomed view of the point cloud map on each region with the same borderline colors as in the left. The blue line represents the RaDiVe odometry estimation. }
    \label{fig:figure_1}
    \vspace{-8mm}
\end{figure}

Data unreliability challenge extends beyond the registration strategy to the selection of reliable points from two key components of scan-to-submap registration: the incoming 4D radar scan and the local submap. {For incoming scans, 4D radar provides precise radial velocity measurements, but previous studies primarily use them as optimization residuals \cite{noh2025garlio,hexel2022dicp}. With limited use of velocity information, point uncertainty modeling largely relies on positional covariance \cite{xu2025incorporating}.} For the local submap, naively accumulating historical 4D radar scans introduces another challenge \cite{li20234d, zhuang20234d, kim2024radar4motion}; noise and outliers accumulate in the local submap, thereby degrading the geometric consistency of the submap. Together, uncertainty in the input 4D radar scan and noise in the local submap compromise odometry accuracy.

To address these challenges in 4D radar odometry, we introduce \textbf{RaDiVe}, a 6-DoF 4D radar odometry framework designed to enable robust \ac{NDT} registration for sparse 4D radar point clouds. RaDiVe mitigates the instability of \ac{NDT} on sparse data and prioritizes reliable voxel matches by restricting correspondences to nearby voxel pairs between the input scan and the local submap. This restriction improves both optimization stability and computational efficiency. To further improve accuracy, we introduce two mechanisms that focus on reliable points from both the input 4D radar scan and the local submap. Specifically, we propose a velocity-discrepancy point uncertainty model to weight input scan points, while employing a \ac{SDF}-based surface point extractor to prune outliers and preserve geometric structure in the local submap. Extensive evaluations across multiple public datasets demonstrate that RaDiVe reduces the average translational \ac{ATE} by $44.4\%$ and rotational \ac{ATE} by $21.3\%$ relative to the {best-performing baselines}, while maintaining real-time performance. \figref{fig:figure_1} shows a representative qualitative odometry and mapping result of RaDiVe and briefly explains the core modules.

Our main contributions are as follows:
\begin{itemize}
    \item \textbf{Distance-Bounded \ac{NDT} Correspondence: } We restrict \ac{NDT} correspondences to nearby voxel pairs between the input scan and the submap, based on the high local overlap between consecutive radar scans. This improves both optimization stability and computational efficiency.
    
    \item \textbf{Velocity-Discrepancy Point Uncertainty: } We propose a point uncertainty model for 4D radar scans based on the discrepancy between measured and propagated radial velocities. The resulting uncertainty provides a reliability measure for each point in the input radar scan.
    
    \item \textbf{\ac{SDF}-based Surface Point Extractor: } We introduce \ac{SDF}-based surface point extraction for the local submap. This mitigates accumulated noise while preserving geometric structure, thereby improving registration accuracy.
    
    \item \textbf{Evaluation on Public Datasets: } RaDiVe is evaluated on multiple public datasets and outperforms previous \ac{SOTA} baselines on most evaluated sequences. We will release our algorithm to the robotics community.    
\end{itemize}

\section{Related Work}
\label{sec:relatedwork}




\subsection{Point Cloud Registration on Radar Odometry}

\subsubsection{ICP-registration in 4D Radar Odometry}


With recent advances in 4D radar point density, subsequent studies incorporate geometric information into odometry estimation by adopting \ac{LiDAR}-inspired point cloud registration, particularly \ac{ICP}. 4DRadarSLAM~\cite{zhang20234dradarslam} employ \ac{GICP} with an adaptive probability distribution. Radar4motion~\cite{kim2024radar4motion} propose polar-grid-based local \ac{RCS} feature extraction for scan-to-submap \ac{ICP}. \citeauthor{herraez2024radar}~\cite{herraez2024radar} introduce Doppler-aware \ac{ICP} for scan-to-scan registration and later incorporated ground constraints to radar-inertial \ac{SLAM} framework RaI-SLAM~\cite{herraez2025ground}. \citeauthor{kim2025doppler}~\cite{kim2025doppler} integrate Doppler-velocity-based correspondences into \ac{ICP}. ClarRO~\cite{furuno2026claro} fuses Doppler and \ac{RCS} to down-weight the radar's weak elevation axis during \ac{ICP} registration. Nevertheless, the reliance of \ac{ICP} on accurate point-level correspondences remains poorly suited to the sparse and low-precision nature of 4D radar. 

\subsubsection{NDT-registration in Radar Odometry}


To address the limitations of \ac{ICP}, \ac{NDT} emerges as a robust alternative. Previous studies demonstrate its effectiveness on 2D \ac{BEV} representations generated by scanning and 4D radars \cite{kung2021normal,zhang2023scan,hilger2024randt}. However, extending \ac{NDT} to radar point clouds remains challenging because the voxel count increases while points per voxel decrease, leading to unreliable voxel statistics and unstable optimization. \citeauthor{li20234d}~\cite{li20234d} demonstrate the potential of distribution-based registration for 4D radar, but they limit registration to keyframes because of point cloud sparsity and computational cost. Thus, achieving stable, real-time \ac{NDT} odometry at the sensor rate remains an open problem. RaDiVe introduces Distance-Bounded \ac{NDT}, which restricts optimization to a reliable set of voxel correspondences, improving robustness while reducing computational cost.

\subsection{Point Reliability in Radar Point Cloud}

Beyond the registration algorithm, selecting reliable points is crucial for registration accuracy. Several methods apply heuristic input filtering, such as an \ac{RCS}-bounded filter \cite{huang2024less}, whereas others estimate point-wise uncertainty based solely on point position \cite{xu2025incorporating} or jointly consolidate point and pose uncertainty \cite{yang2026geometrically}. Although the importance of radial velocity is recognized, it is generally incorporated as an additional residual term in optimization \cite{noh2025garlio, hexel2022dicp}, rather than a point-reliability term. Consequently, incorporating radial velocity measurements to derive uncertainty weight for each point remains insufficiently explored. RaDiVe addresses this by introducing a radial-velocity-informed point uncertainty model.


\section{Method}
\label{sec:method}

\begin{figure}[!t]
    \centering
    \includegraphics[width=0.9\columnwidth]{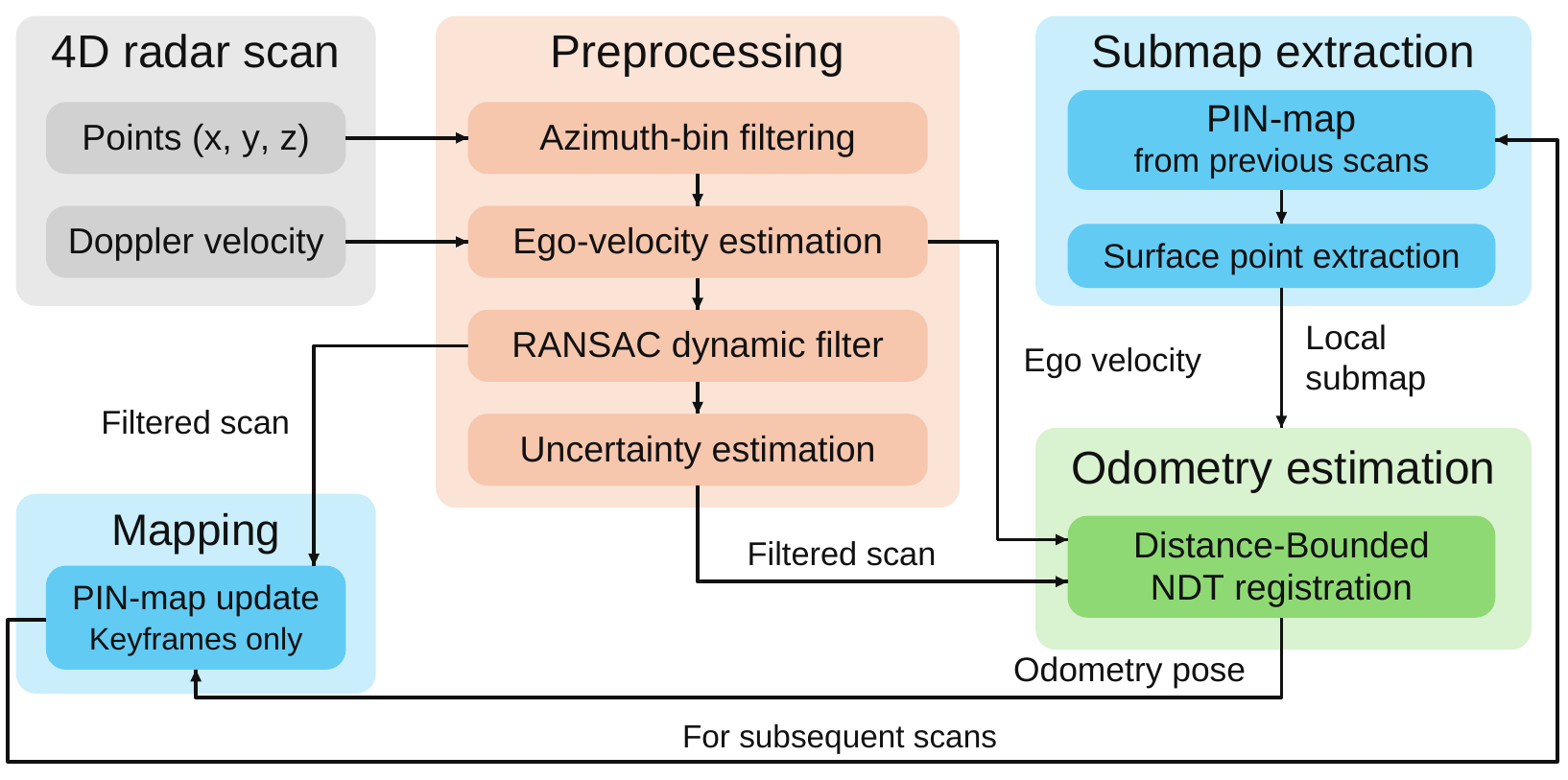}
    \vspace{-4mm}
    \caption{Overall pipeline of RaDiVe. }
    \label{fig:pipeline}
    \vspace{-6mm}
\end{figure}


We first preprocess to remove dynamic points and duplicate returns caused by radio-wave penetration. After extracting reliable points from the input radar scan, we estimate the ego-velocity and calculate point uncertainty using radial-velocity measurements. {Surface point extraction module constructs a local submap from the global map. The current scan is then aligned with this submap using Distance-Bounded \ac{NDT} registration. After pose estimation, selected keyframes are used to update the map.} \figref{fig:pipeline} presents the pipeline overview.

\subsection{Radar Pointcloud Preprocessing}
\label{subsec:preprocessing}
The raw radar scan is preprocessed to remove noise, dynamic returns, and duplicated points before registration. For input 4D radar pointcloud  $\mathcal{P} = \{(x_k, y_k, z_k, v_k)$, where $(x,y,z)\in\mathbb{R}^3,\;v\!=\!\text{Radial Velocity} \}_{k=1}^N$, we apply the following preprocessing steps to retain reliable static points. 

\subsubsection{Azimuth-bin filtering}
One notable difference between \ac{LiDAR} and 4D radar is that 4D radar scans contain duplicated returns along the same direction due to radio-wave penetration. Points generated from scattering and reflection when penetrating objects may be inaccurate; therefore, we integrate azimuth-bin filtering. We start by computing the range, azimuth, and elevation ($r_k, \alpha_k, \varepsilon_k$) of each point as follows:
\begingroup
\edef\savedbodybaselineskip{\the\baselineskip}
\small
\setlength{\baselineskip}{\savedbodybaselineskip}
\addtolength{\abovedisplayskip}{-2mm}
\addtolength{\belowdisplayskip}{-2mm}
\addtolength{\abovedisplayshortskip}{-2mm}
\addtolength{\belowdisplayshortskip}{-2mm}
\begin{align}
r_k &= \sqrt{x_k^2 + y_k^2 + z_k^2}, \\
\alpha_k &= \arctan(y_k, x_k), \quad \alpha_k \in [-\pi, \pi], \\
\varepsilon_k &= \arcsin\!\bigl(z_k/ r_k \bigr), \quad \varepsilon_k \in [-\tfrac{\pi}{2}, \tfrac{\pi}{2}].
\end{align}
\endgroup

Since many real-world environments contain meaningful structural information in the \ac{BEV}, which primarily represents the azimuthal structure \cite{luo2023bevplace}, we quantize only the azimuth direction while preserving the original elevation values. We quantize the azimuth into discrete bins of width $\Delta\alpha$:
\begingroup
\edef\savedbodybaselineskip{\the\baselineskip}
\small
\setlength{\baselineskip}{\savedbodybaselineskip}
\addtolength{\abovedisplayskip}{-2mm}
\addtolength{\belowdisplayskip}{-2mm}
\addtolength{\abovedisplayshortskip}{-2mm}
\addtolength{\belowdisplayshortskip}{-2mm}
\begin{align}
b_k &= \left\lfloor \frac{\alpha_k + \pi}{\Delta\alpha} \right\rfloor, 
\quad b_k \in \{0,1,\dots,B-1\},\\
e_k &= \varepsilon_k . 
\end{align}
\endgroup
{We group points by azimuth bin and elevation. Within each group $(b,e)$, we retain the closest point as follows:}
\begingroup
\edef\savedbodybaselineskip{\the\baselineskip}
\small
\setlength{\baselineskip}{\savedbodybaselineskip}
\addtolength{\abovedisplayskip}{-2mm}
\addtolength{\belowdisplayskip}{-2mm}
\addtolength{\abovedisplayshortskip}{-2mm}
\addtolength{\belowdisplayshortskip}{-2mm}
\begin{equation}
k^*_{b,e} = \underset{k:\,(b_k,e_k)=(b,e)}{\arg\min}\;r_k.
\end{equation}
\endgroup
{Azimuth-bin filtering serves as an auxiliary preprocessing step intended to suppress potential duplicated returns.}

\subsubsection{Dynamic point filtering via RANSAC and ego-velocity}
\label{subsubsec:ransac}
To filter out dynamic points and calculate robust ego-velocity, we employ \ac{RANSAC}-based filtering and \ac{LSQ}-based ego-velocity calculation similar to existing methods \cite{kellner2013instantaneous, kellner2014instantaneous, park20213d, jung2024co}. For a return generated by a static object, its radial velocity should equal the projection of the negative sensor ego-velocity onto the point direction. By extending sinusoidal fitting from 2D to 3D, the 3D ego-velocity $\mathbf{v}$ of the robot and static points are extracted. 

\subsection{Velocity-Discrepancy Point Uncertainty}
\label{subsec:unc}
{Existing approaches model radar positional uncertainty in polar coordinates \cite{xu2025incorporating}. Doppler measurements provide complementary information for radar tracking \cite{jacobs2023vtc}. Our velocity-discrepancy point uncertainty represents a Doppler-consistency-based reliability cue for registration, rather than a complete model of 3D positional uncertainty.}

\begin{figure}[!t]
    \centering
    \includegraphics[width=0.5\columnwidth]{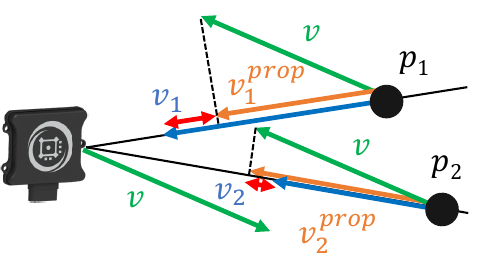}
    \vspace{-4mm}
    \caption{Visualization of velocity-discrepancy point uncertainty. Blue and orange arrows represent Doppler measured and ego velocity-propagated radial velocity, respectively. The larger discrepancy at $p_1$ indicates higher uncertainty and a lower reliability weight. }
    \label{fig:uncertainty}
    \vspace{-5mm}
\end{figure}

An overview of the proposed velocity-discrepancy point uncertainty is shown in \figref{fig:uncertainty}. Given the robust ego-velocity $\mathbf{v}$ from \ref{subsubsec:ransac}, we project it to each static point. For a static point $p_k$, its unit direction vector is $ \mathbf{d}_k = \frac{(x_k,y_k,z_k)^\top}{\|\,(x_k,y_k,z_k)\|\!}$. The propagated radial velocity $v_k^{prop}$ is calculated as $v_k^\text{prop} = - \mathbf{d}_k^\top v$. 

We calculate the discrepancy between the measured and propagated radial velocities as $\Delta v_k = \bigl|\,v_k^\text{prop} - v_k\bigr|$. Because $\Delta v_k \in [0, \infty )$, we bound the reliability weight $w_k \in (0, 1]$ using an exponential function as follows: 
\begingroup
\edef\savedbodybaselineskip{\the\baselineskip}
\small
\setlength{\baselineskip}{\savedbodybaselineskip}
\addtolength{\abovedisplayskip}{-2mm}
\addtolength{\belowdisplayskip}{-2mm}
\addtolength{\abovedisplayshortskip}{-2mm}
\addtolength{\belowdisplayshortskip}{-2mm}
\begin{equation}
w_k
= \exp\bigl(- \Delta v_k\bigr)
= \exp\bigl(-\,\lvert v_k^\text{prop} - v_k\rvert\bigr).
\label{eq:unc}
\end{equation}
\endgroup

{Radial velocities are expressed in $\mathrm{m/s}$, while \eqref{eq:unc} uses their numerical values, corresponding to a velocity scale of $1\,\mathrm{m/s}$. For fixed measured Doppler velocity and ego-velocity, the weight is insensitive to displacement along the viewing ray. Ego-velocity estimation errors can also affect its reliability.}

\subsection{Surface Point Extraction for Submap}
To construct a low-noise local map for scan-to-submap registration in \ref{subsec:NDT}, we extract points lying on stable surfaces. We leverage the {\ac{PIN}}-mapping~\cite{pan2024pin} for its fast, robust mapping based on an implicit neural representation that learns an \ac{SDF} from input point clouds. \ac{PIN}-map represents local geometry using neural points, each defined by a position $x_k\in\mathbb{R}^3$, orientation $q_k\in\mathrm{SO}(3)$, and latent geometric feature $f^g_k\in\mathbb{R}^d$.
{Following PIN-SLAM~\cite{pan2024pin}, the implicit \ac{SDF} decoder $D_\theta$ is trained using samples along measurement rays, including measured endpoints, Gaussian samples around them, and uniformly sampled points in front of and behind them. For a measured range $\rho$ and a sample distance $\ell$ along the same ray, the supervision target is the signed projective distance $\rho-\ell$, which is positive before the return and negative behind it. Training uses sigmoid-based binary cross-entropy with Eikonal regularization, while we retain this sampling and supervision formulation.}

{Registration is performed for every scan, while the map is updated at selected keyframes after pose estimation. A keyframe is selected when the translation or yaw change relative to the previous keyframe exceeds $\delta_t$ or $\delta_r$, respectively. Representative settings are $\delta_t=2\meter$ and $\delta_r=3\degree$, while the specific thresholds may vary across sequences. The local neural-point subset is refreshed for each scan based on distance to the sensor and the travel distance since each point was created, while excluded points remain in the global \ac{PIN}-map.}

For submap generation, rather than generating candidate points, we directly use each neural point position in the trained \ac{PIN}-map as a query to evaluate its learned \ac{SDF}. \ac{SDF} of each neural point $x_j$ is computed using its neighboring neural points $x_k$ as $ s_{jk} = D_\theta\bigl(f^g_k,\;d_{jk}\bigr)$, where $d_{jk} = q_k (x_j - x_k) q_k^{-1}$. 
These predictions are aggregated through inverse-distance weighting to emphasize nearby neural points:
\begingroup
\edef\savedbodybaselineskip{\the\baselineskip}
\small
\setlength{\baselineskip}{\savedbodybaselineskip}
\addtolength{\abovedisplayskip}{-2mm}
\addtolength{\belowdisplayskip}{-2mm}
\addtolength{\abovedisplayshortskip}{-2mm}
\addtolength{\belowdisplayshortskip}{-2mm}
\begin{align}
    S(x_i) &= \frac{\sum_{j \in \mathcal{N}(x_i)} \nu_{ij} s_{ij}}{\sum_{j \in \mathcal{N}(x_i)} \nu_{ij}}, \quad\text{with}\quad \nu_{ij} = \| x_i - x_j \|^{-2},
\end{align}
\endgroup
where \(\mathcal{N}(x_i)\) is \(K\)-nearest neighbors found via voxel hashing.

The resulting \ac{SDF} values allow us to extract reliable surface points by retaining points near the implicit surface. 
{We empirically set the surface threshold to $\tau_s=0.01\meter$ to balance surface selectivity and point availability for registration.}
\begingroup
\edef\savedbodybaselineskip{\the\baselineskip}
\small
\setlength{\baselineskip}{\savedbodybaselineskip}
\addtolength{\abovedisplayskip}{-2mm}
\addtolength{\belowdisplayskip}{-2mm}
\addtolength{\abovedisplayshortskip}{-2mm}
\addtolength{\belowdisplayshortskip}{-2mm}
\begin{equation}
\mathcal{P}_{\mathrm{surf}}
= \bigl\{\,x_i \in \mathcal{M}\;\bigm|\; |S(x_i)| \le \tau_s \bigr\},
\label{eq:surface_set2}
\end{equation}
\endgroup
where \(\mathcal{M}\) denotes the set of neural point positions in the local \ac{PIN}-map. $\mathcal{P}_{\mathrm{surf}}$ provides a noise-filtered and geometrically consistent surface representation for \ac{NDT}-based registration.

\subsection{Distance-Bounded NDT for Radar Odometry}
\label{subsec:NDT}

We denote the filtered current radar scan as the source point cloud and the extracted surface points $\mathcal{P}_{\mathrm{surf}}$ as the target submap. We perform \ac{NDT} registration using the reliability weights derived from the velocity-discrepancy point uncertainty. {For voxel-based Gaussian construction, each voxel $V_i$ containing $n_i$ points is characterized by mean $\mu_i$, covariance $\Sigma_i$, and average weight $w_i$:}
\begingroup
\edef\savedbodybaselineskip{\the\baselineskip}
\small
\setlength{\baselineskip}{\savedbodybaselineskip}
\addtolength{\abovedisplayskip}{-2mm}
\addtolength{\belowdisplayskip}{-2mm}
\addtolength{\abovedisplayshortskip}{-2mm}
\addtolength{\belowdisplayshortskip}{-2mm}
\begin{align}
\Sigma_i &= \frac{1}{n_i}\sum_{j=1}^{n_i}(p_j - \mu_i)(p_j - \mu_i)^\top + \epsilon I, \\
\mu_i &= \frac{1}{n_i}\sum_{j=1}^{n_i}p_j, ~~~~ w_i = \frac{1}{n_i}\sum_{j=1}^{n_i} w_{p_j},
\end{align}
\endgroup
where $p_j$ denotes a point within voxel $V_i$, $w_{p_j}$ is velocity-discrepancy point uncertainty weight from \ref{subsec:unc}, and {$\epsilon=0.01\,\mathrm{m}^{2}$ is the covariance regularization constant.}
We use anisotropic voxels of size $2\meter\times2\meter\times1\meter$ to provide sufficient spatial support. {Source-point weights are averaged to scale the \ac{NDT} matching cost, while voxel means and covariances remain unweighted. This reduces the influence of Doppler-inconsistent source Gaussians without changing their shape.}

\paragraph{Correspondence Selection}

{We construct source Gaussians $(\mu_A,\Sigma_A,w)$ in the current radar frame and target Gaussians $(\mu_B,\Sigma_B)$ after expressing the submap in the previous radar frame. Before applying the initial relative transformation, we retain all pairs satisfying $\|\mu_A-\mu_B\|<d_{\max}$, with a representative threshold of $d_{\max}=10\meter$. These correspondences remain fixed throughout LM optimization. Restricting the pair set excludes distant matches and reduces the number of pairs evaluated during optimization.}


\paragraph{Cost Function}
The transformation between consecutive radar frames is parameterized as a 6-\ac{DoF} vector \(\xi = [r, p, y, t_x, t_y, t_z]^\top\), where \((r, p, y)\) denote roll, pitch, yaw, and \((t_x, t_y, t_z)\) represent translations. The optimal transformation $\xi^*$ minimizes the following weighted \ac{NDT} cost function as:
\begingroup
\edef\savedbodybaselineskip{\the\baselineskip}
\small
\setlength{\baselineskip}{\savedbodybaselineskip}
\addtolength{\abovedisplayskip}{-2mm}
\addtolength{\belowdisplayskip}{-2mm}
\addtolength{\abovedisplayshortskip}{-2mm}
\addtolength{\belowdisplayshortskip}{-2mm}
\begin{align}
\label{eq:dist_function}
Cost(\xi) &= \sum_{i}\left[1 - \frac{\exp\left(-\frac{1}{2}\Delta_i(\xi)^\top \Sigma_i(\xi)^{-1}\Delta_i(\xi)\right)}{\sqrt{(2\pi)^3|\Sigma_i(\xi)|}}\right] w_i,
\end{align}
\endgroup
where the difference and covariance are defined as
\begingroup
\edef\savedbodybaselineskip{\the\baselineskip}
\small
\setlength{\baselineskip}{\savedbodybaselineskip}
\addtolength{\abovedisplayskip}{-2mm}
\addtolength{\belowdisplayskip}{-2mm}
\addtolength{\abovedisplayshortskip}{-2mm}
\addtolength{\belowdisplayshortskip}{-2mm}
\begin{align}
\Delta_i(\xi) &= R(\xi)\mu_{A,i} + t(\xi) - \mu_{B,i}, \\[4pt]
\Sigma_i(\xi) &= R(\xi)\Sigma_{A,i}R(\xi)^\top + \Sigma_{B,i}.
\end{align}
\endgroup
Here, $R(\xi)$ and $t(\xi)$ are the rotation matrix and translation vector derived from Euler angles and translations, respectively. {We use this local Euler angle parameterization for the small inter-scan rotations expected in the ground vehicle setting.}

\paragraph{Ego-Velocity Translation Prior}
To improve robustness and convergence for low-speed motion, we incorporate a translation prior using the ego-velocity estimate in \ref{subsubsec:ransac}:
\begingroup
\edef\savedbodybaselineskip{\the\baselineskip}
\small
\setlength{\baselineskip}{\savedbodybaselineskip}
\addtolength{\abovedisplayskip}{-2mm}
\addtolength{\belowdisplayskip}{-2mm}
\addtolength{\abovedisplayshortskip}{-2mm}
\addtolength{\belowdisplayshortskip}{-2mm}
\begin{align}
Cost_{\text{prior}}(\xi)=\frac{\lambda_{\text{trans}}}{2}\|\mathbf{t}(\xi)-\mathbf{t}_{\text{init}}\|^2,
\end{align}
\endgroup
where \(\mathbf{t}_{\text{init}}\) denotes the initial translation from ego-motion, activated when the ego-velocity is lower than $1.5 \meter / \sec$.

\paragraph{\ac{LM} Optimization}
We employ the \ac{LM} algorithm to iteratively update the transformation parameters $\xi$ by solving $(\mathbf{H} + \lambda\,\mathrm{diag}(\mathbf{H}))\,\delta\xi = -\mathbf{g}$. Upon convergence, the optimal transformation is given by:
\begingroup
\edef\savedbodybaselineskip{\the\baselineskip}
\small
\setlength{\baselineskip}{\savedbodybaselineskip}
\addtolength{\abovedisplayskip}{-2mm}
\addtolength{\belowdisplayskip}{-2mm}
\addtolength{\abovedisplayshortskip}{-2mm}
\addtolength{\belowdisplayshortskip}{-2mm}
\begin{align}
T = \begin{bmatrix}
R(\xi^*) & t(\xi^*)\\[4pt]
0 & 1
\end{bmatrix}.
\end{align}
\endgroup

{LM is initialized using ego-velocity and the previous inter-scan rotation. We use at most five iterations, terminating when an update decreases the cost by less than $10^{-6}$.}



\section{Experiments}
\label{sec:experiment}

\begin{table*}[t]
\centering
\caption{Evaluation Results in \texttt{Ntu4dradlm} with Oculii}
\vspace{-2mm}
\label{tab:NTU_exp}
\resizebox{0.95\textwidth}{!}{
\begin{tabular}{c|cccc|cccc|cccc|cccc}
\toprule
 & \multicolumn{4}{c|}{\texttt{loop3}} & \multicolumn{4}{c|}{\texttt{loop2}} & \multicolumn{4}{c|}{\texttt{cp}} & \multicolumn{4}{c}{\texttt{nyl}} \\ 
 & ATE$_t$ & ATE$_r$ & {RTE$_t$} & {RTE$_r$} & ATE$_t$ & ATE$_r$ & {RTE$_t$} & {RTE$_r$} & ATE$_t$ & ATE$_r$ & {RTE$_t$} & {RTE$_r$} & ATE$_t$ & ATE$_r$ & {RTE$_t$} & {RTE$_r$} \\ \midrule
\textbf{4DRadarSLAM} & \underline{12.802} & 2.074 & {2.166} & {0.00208} & \textbf{54.733} & \underline{5.954} & {5.349} & {\underline{0.00412}} & \underline{1.852} & \underline{1.189} & {3.644} & {\underline{0.01672}} & {6.253} & \underline{1.477} & {\underline{3.279}} & {0.00752} \\ 
\textbf{Radar4Motion} & 91.364 & 7.685 & {4.553} & {0.00485} & 255.422 & 17.837 & {11.504} & {0.01367} & 2.710 & 2.051 & {4.823} & {0.02309} & {\underline{5.397}} & 3.082 & {4.555} & {0.00742} \\ 
\textbf{KISS-ICP} & 17.297 & \underline{1.594} & {\underline{1.807}} & {\underline{0.00181}} & 85.657 & 6.937 & {\underline{4.536}} & {0.00474} & 2.980 & 2.493 & {5.086} & {0.02176} & 6.952 & 2.653 & {4.719} & {\underline{0.00690}} \\ 
\textbf{Doppler-ICP} & 29.897 & 2.287 & {2.567} & {0.00250} & 133.028 & 10.658 & {7.085} & {0.00738} & 2.304 & 2.393 & {\underline{3.602}} & {0.02355} & 14.552 & 6.507 & {5.780} & {0.02087} \\ \midrule
\textbf{Ours} & \textbf{9.914} & \textbf{1.147} & {\textbf{1.396}} & {\textbf{0.00151}} & \underline{70.825} & \textbf{5.928} & {\textbf{3.451}} & {\textbf{0.00396}} & \textbf{0.387} & \textbf{0.768} & {\textbf{1.179}} & {\textbf{0.01012}} & \textbf{2.970} & \textbf{1.472} & {\textbf{1.888}} & {\textbf{0.00364}} \\ \bottomrule
\end{tabular}}
\vspace{-2mm}
\end{table*}

\begin{table*}[t]
\centering
\vspace{-2mm}
\caption{Evaluation Results in \texttt{HeRCULES} with ARS548}
\vspace{-2mm}
\label{tab:Hercules_exp}
\resizebox{0.95\textwidth}{!}{
\begin{tabular}{c|cccc|cccc|cccc|cccc}
\toprule
 & \multicolumn{4}{c|}{\texttt{Library}} & \multicolumn{4}{c|}{\texttt{Sports Complex}} & \multicolumn{4}{c|}{\texttt{Parking Lot}} & \multicolumn{4}{c}{\texttt{Street}} \\ 
 & ATE$_t$ & ATE$_r$ & {RTE$_t$} & {RTE$_r$} & ATE$_t$ & ATE$_r$ & {RTE$_t$} & {RTE$_r$} & ATE$_t$ & ATE$_r$ & {RTE$_t$} & {RTE$_r$} & ATE$_t$ & ATE$_r$ & {RTE$_t$} & {RTE$_r$} \\ \midrule
\textbf{4DRadarSLAM} & 120.454 & 21.888 & {28.690} & {0.07228} & 108.178 & 59.607 & {29.304} & {0.15423} & 17.173 & 13.974 & {16.308} & {0.08448} & 151.787 & 61.270 & {91.328} & {0.44790} \\ 
\textbf{Radar4Motion} & 8.311 & 4.149 & {1.808} & {0.00894} & 3.899 & \underline{1.752} & {1.760} & {\underline{0.00447}} & 1.712 & \underline{2.642} & {2.744} & {\underline{0.01415}} & 5.463 & \underline{4.582} & {4.585} & {0.01475} \\ 
\textbf{KISS-ICP} & \underline{5.347} & \underline{3.365} & {\underline{1.281}} & {\underline{0.00731}} & \underline{2.843} & 2.004 & {\underline{1.715}} & {0.00582} & \underline{1.329} & 2.959 & {\underline{2.683}} & {\textbf{0.01045}} & \underline{4.154} & 5.119 & {\textbf{3.730}} & {\underline{0.00937}} \\ 
\textbf{Doppler-ICP} & - & - & {-} & {-} & - & - & {-} & {-} & - & - & {-} & {-} & - & - & {-} & {-} \\ \midrule
\textbf{Ours} & \textbf{2.179} & \textbf{2.901} & {\textbf{0.927}} & {\textbf{0.00333}} & \textbf{1.691} & \textbf{1.484} & {\textbf{1.555}} & {\textbf{0.00366}} & \textbf{1.034} & \textbf{1.026} & {\textbf{2.484}} & {0.01728} & \textbf{2.600} & \textbf{2.571} & {\underline{3.743}} & {\textbf{0.00466}} \\ \bottomrule
\end{tabular}}
\vspace{-5mm}
\end{table*}

\begin{figure*}[!t]
    \centering
    \includegraphics[width=0.95\textwidth]{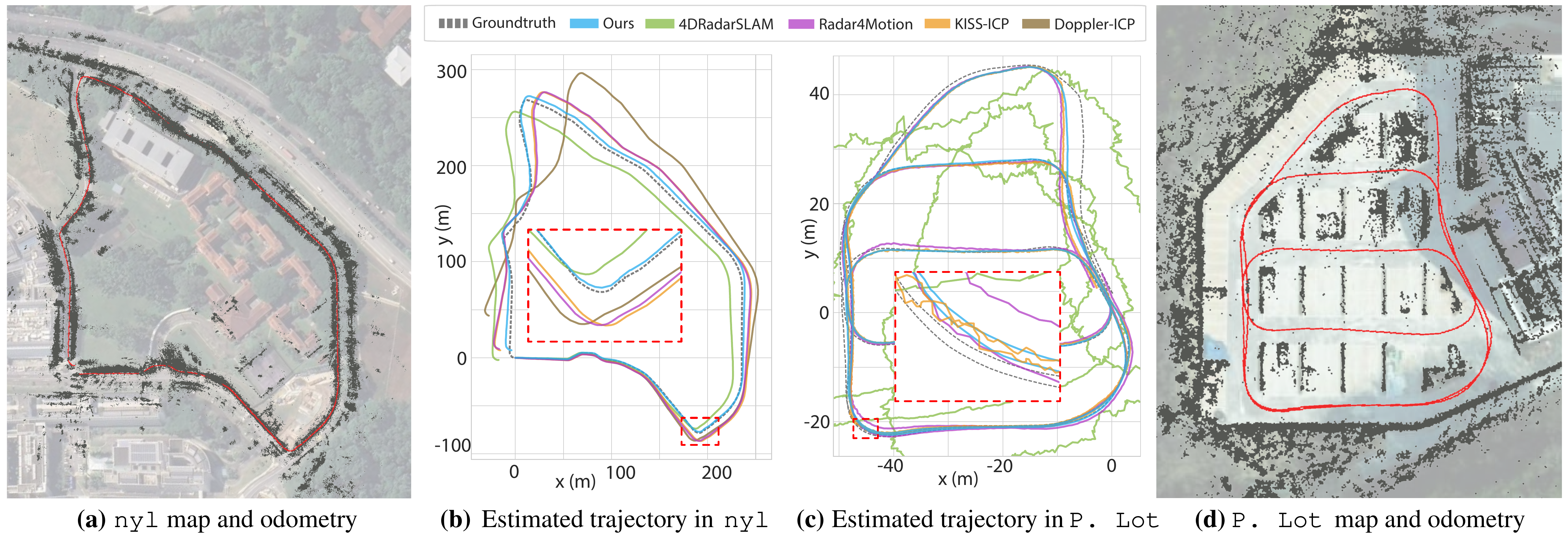}
    \vspace{-4mm}
    \caption{Qualitative analysis of \texttt{nyl} and \texttt{Parking Lot Night 3}. Red lines in Fig. 4a and Fig. 4d are estimated odometry.}
    \vspace{-6mm}
\end{figure*}

\subsection{Dataset and Evaluation}

Our approach is validated on three public 4D radar benchmarks: \texttt{Ntu4dradlm}~\cite{zhang2023ntu4dradlm}, \texttt{HeRCULES}~\cite{kim2025hercules}, and \texttt{SNAIL}~\cite{huai2024snail}. \texttt{Ntu4dradlm} provides Oculii Eagle scans on both a car and a handcart platform; \texttt{HeRCULES} uses a car‐mounted Continental ARS548 over multiple loops or a traffic-jammed scenario; \texttt{SNAIL} contains sequences from both sensors attached to a car-based system. 
Oculii Eagle yields $\sim$7500 points/frame at 12 Hz, whereas Continental ARS548 returns $\sim$150 points/frame at 20 Hz, differing heavily in sensor frequency, point throughput, and even point precision. {Azimuth-bin filtering is enabled for Oculii Eagle with $\Delta\alpha=0.175^\circ$, while disabled for Continental ARS548 to avoid further reducing the already sparse input.} {To accommodate the extreme sparsity of Continental ARS548, each source point is represented by a Gaussian centered at that point with fixed isotropic covariance $\Sigma_i=(0.25\meter)^2I$. Target Gaussians remain voxel-based, with covariances computed using the formulation in \ref{subsec:NDT}, while the registration formulation remains unchanged.}

Odometry accuracy is measured with the Evo evaluator~\cite{grupp2017evo} via \ac{RMSE} of translational [\meter] and rotational \ac{ATE} [\degree] to evaluate the accuracy and stability of methods. {Estimated and ground-truth trajectories are associated by nearest timestamps, rigidly aligned in SE(3) without scale correction, and then projected onto the XY plane. We report the RMSE of translational error as ATE$_t$ [\meter] and rotational error as ATE$_r$ [\degree].}
{We additionally report translational [\%] and rotational [\degree/\meter] \ac{RTE}, averaged over all evaluated sub-trajectories spanning 10--50\% of each sequence's ground-truth path length~\cite{zhang2018tutorial}.}
We benchmark against open‐source \ac{ICP} based 4D radar‐only odometry methods: 4DRadarSLAM~\cite{zhang20234dradarslam}, Radar4Motion~\cite{kim2024radar4motion}. Furthermore, to compare with recent \ac{ICP} algorithms, KISS-ICP~\cite{vizzo2023kiss} and Doppler-ICP~\cite{hexel2022dicp} are chosen for their generality and explicit design for 4D point clouds. 
{We report the quantitative evaluation of RaDiVe to horizontal translation and yaw.}
{For a fair odometry comparison, loop closure is disabled in all baseline methods.} In the result tables, best results are in \textbf{Bold}, the second best results are \underline{Underlined}, and ‘–’ indicates convergence failure during optimization.  

\subsection{Evaluation on the \texttt{Ntu4dradlm} Dataset}

As \tabref{tab:NTU_exp}, RaDiVe achieves the lowest \ac{ATE}$_t$ and \ac{ATE}$_r$ across most sequences in \texttt{Ntu4dradlm}, outperforming point correspondence–based methods by a wide margin. 
In particular, for shaky handcart sequences (\texttt{cp}, \texttt{nyl}), RaDiVe demonstrates strong robustness from the combination of \ac{NDT}-based registration and a translation prior, with the \ac{ATE}$_t$ in \texttt{cp} being about $5\times$ lower than any baseline. 
Even on smoother car-platform sequences (\texttt{loop2}, \texttt{loop3}), RaDiVe maintains superior accuracy, highlighting its ability to preserve global consistency. 
4DRadarSLAM achieves the second-best result in most cases and even outperforms RaDiVe on \texttt{loop2} with \ac{ATE}$_t$ for its compatibility with \texttt{Ntu4dradlm}. 
{RaDiVe also achieves the lowest translational and rotational \ac{RTE} on all sequences. Notably, on \texttt{loop2}, RaDiVe yields lower relative errors than 4DRadarSLAM despite its higher translational \ac{ATE}.}
Qualitative results are visualized in Fig. 4a and Fig. 4b for the \texttt{nyl} sequence, where other methods suffer from global drift while RaDiVe robustly estimates odometry. 

\subsection{Evaluation on the \texttt{HeRCULES} Dataset}

\tabref{tab:Hercules_exp} shows that RaDiVe substantially outperforms other radar-only baselines on both \ac{ATE}$_t$ and \ac{ATE}$_r$ {across} every sequence. 
This is primarily due to the distance-bounded \ac{NDT} and velocity-discrepancy point uncertainty, which are particularly effective in the highly sparse but precise radial velocity of Continental ARS548, leading to accurate odometry estimation. 
\texttt{Street Day 1} introduces additional challenges: traffic congestion and stop-and-go motion in the urban environment, which degrade data quality compared to open-area sequences. 
While 4DRadarSLAM and Doppler-ICP fail under these conditions, RaDiVe maintains reliable odometry due to its translation prior scheme based on ego-velocity. {For relative errors, RaDiVe achieves the lowest translational and rotational \ac{RTE} on three of the four sequences for each metric.}

Fig. 4c and Fig. 4d include odometry and mapping results of RaDiVe in \texttt{Parking Lot Night 3}, confirming its robustness and stability. 
4DRadarSLAM and Doppler-ICP diverge due to registration failure. KISS-ICP exhibits high local jittering, as evident in the zoomed view, resulting from the registration instability of ICP on the extremely sparse 4D radar point cloud of Continental ARS548. 
Although Radar4Motion presents the only smooth trajectory among the comparison baselines, it suffers from global drift, resulting in a different trajectory on every loop. 
In comparison, RaDiVe successfully estimated a smooth and accurate trajectory due to its distance-bounded \ac{NDT} and velocity-discrepancy uncertainty.

\begin{table*}[t]
\centering
\caption{Evaluation Results in \texttt{SNAIL} with ARS548}
\vspace{-2mm}
\label{tab:SNAIL_ars_exp}
\resizebox{0.95\textwidth}{!}{
\begin{tabular}{c|cccc|cccc|cccc}
\toprule
 & \multicolumn{4}{c|}{\texttt{20240113\_1}} & \multicolumn{4}{c|}{\texttt{20240113\_3}}  & \multicolumn{4}{c}{\texttt{20240123\_3}} \\ 
 & ATE$_t$ & ATE$_r$ & {RTE$_t$} & {RTE$_r$} & ATE$_t$ & ATE$_r$ & {RTE$_t$} & {RTE$_r$} & ATE$_t$ & ATE$_r$ & {RTE$_t$} & {RTE$_r$} \\ \midrule
\textbf{4DRadarSLAM} & 21.602 & 3.617 & {15.950} & {0.03530} & 134.225 & 8.739 & {13.897} & {0.00826} & 264.611 & 25.883 & {26.134} & {0.04492} \\ 
\textbf{Radar4Motion} & 10.065 & 10.735 & {5.800} & {0.05072} & 72.672 & 10.342 & {7.189} & {0.00738} & 22.974 & 9.351 & {4.443} & {0.01107} \\ 
\textbf{KISS-ICP} & \underline{3.139} & \underline{3.393} & {\underline{2.752}} & {\underline{0.01580}} & \underline{25.282} & \textbf{3.064} & {\textbf{2.537}} & {\underline{0.00196}} & \underline{5.166} & \textbf{3.030} & {\underline{2.884}} & {\underline{0.00510}} \\ 
\textbf{Doppler-ICP} & - & - & {-} & {-} & - & - & {-} & {-} & - & - & {-} & {-} \\ \midrule
\textbf{Ours} & \textbf{0.885} & \textbf{0.963} & {\textbf{2.037}} & {\textbf{0.00780}} & \textbf{3.521} & \underline{4.782} & {\underline{2.870}} & {\textbf{0.00118}} & \textbf{2.480} & \underline{5.554} & {\textbf{2.403}} & {\textbf{0.00191}} \\ \bottomrule
\end{tabular}}
\vspace{-2mm}
\end{table*}

\begin{table*}[t]
\centering
\vspace{-2mm}
\caption{Evaluation Results in \texttt{SNAIL} with Oculii Eagle}
\vspace{-2mm}
\label{tab:SNAIL_oculii_exp}
\resizebox{0.95\textwidth}{!}{
\begin{tabular}{c|cccc|cccc|cccc}
\toprule
 & \multicolumn{4}{c|}{\texttt{20240113\_1}} & \multicolumn{4}{c|}{\texttt{20240113\_3}}  & \multicolumn{4}{c}{\texttt{20240123\_3}} \\ 
 & ATE$_t$ & ATE$_r$ & {RTE$_t$} & {RTE$_r$} & ATE$_t$ & ATE$_r$ & {RTE$_t$} & {RTE$_r$} & ATE$_t$ & ATE$_r$ & {RTE$_t$} & {RTE$_r$} \\ \midrule
\textbf{4DRadarSLAM} & \underline{10.750} & 60.474 & {\underline{8.821}} & {\underline{0.07664}} & 116.532 & 88.049 & {7.381} & {0.01241} & \underline{64.069} & 18.662 & {11.436} & {\underline{0.02108}} \\ 
\textbf{Radar4Motion} & 25.690 & 33.468 & {12.660} & {0.17292} & 88.049 & 45.855 & {8.579} & {0.02507} & 79.615 & \underline{17.896} & {\underline{11.312}} & {0.03661} \\ 
\textbf{KISS-ICP} & 19.460 & \underline{24.796} & {13.808} & {0.12205} & 57.273 & 34.600 & {\underline{5.226}} & {\textbf{0.00974}} & 87.179 & 19.811 & {15.122} & {0.04265} \\ 
\textbf{Doppler-ICP} & 40.944 & 54.793 & {19.596} & {0.26977} & \textbf{34.600} & \underline{19.370} & {5.307} & {0.01204} & 295.243 & 44.363 & {26.529} & {0.07998} \\ \midrule
\textbf{Ours} & \textbf{1.497} & \textbf{1.385} & {\textbf{1.195}} & {\textbf{0.00923}} & \underline{41.886} & \textbf{5.459} & {\textbf{4.861}} & {\underline{0.01050}} & \textbf{18.912} & \textbf{17.668} & {\textbf{2.671}} & {\textbf{0.00748}} \\ \bottomrule
\end{tabular}}
\vspace{-5mm}
\end{table*}

\begin{figure*}[!t]
    \centering
    \includegraphics[width=0.95\textwidth]{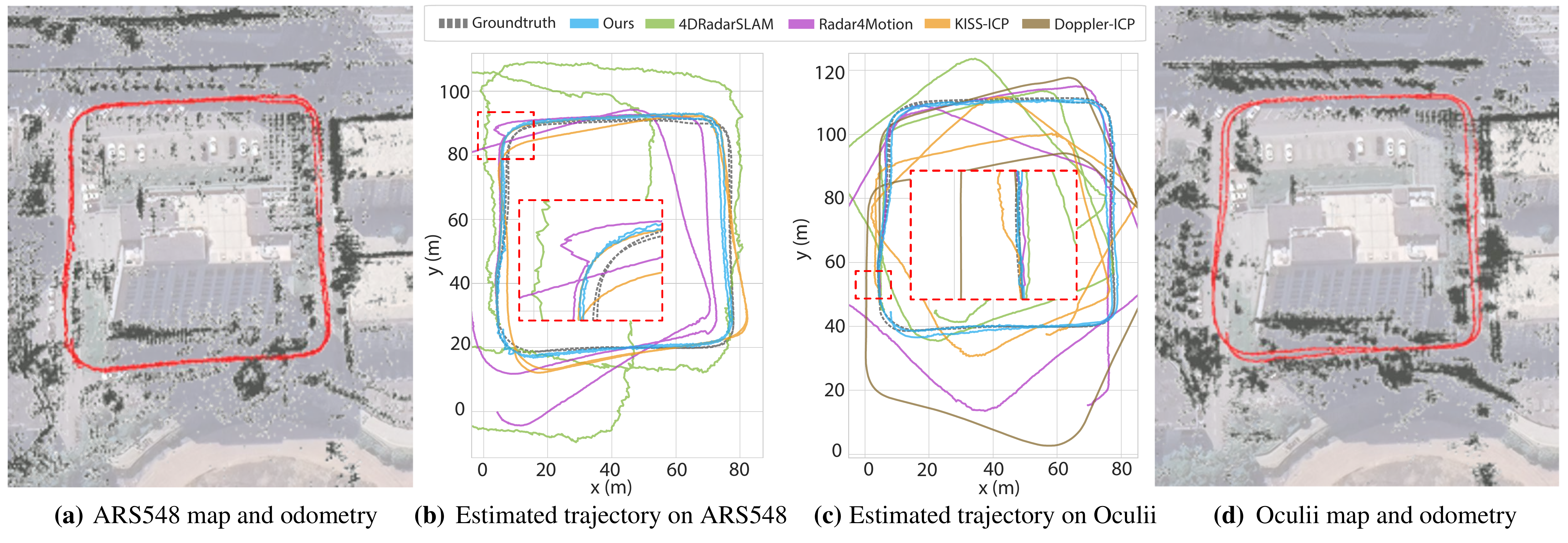}
    \vspace{-4mm}
    \caption{Qualitative result of \texttt{20240113\_1} on Continental ARS548 and Oculii Eagle. Red lines in Fig. 5a and Fig. 5d are estimated odometry.}
    \vspace{-6mm}
\end{figure*}

\subsection{Evaluation on the \texttt{SNAIL} Dataset}

\tabref{tab:SNAIL_ars_exp} and \tabref{tab:SNAIL_oculii_exp} summarize results on the \texttt{SNAIL} dataset for Continental ARS548 and Oculii Eagle, respectively. 
On Continental ARS548, RaDiVe consistently outperforms radar-only baselines in \ac{ATE}$_t$ and maintains competitive \ac{ATE}$_r$. 
Similar to \texttt{HeRCULES}, Doppler-ICP fails to converge due to the extreme sparsity of the Continental ARS548, while 4DRadarSLAM and Radar4Motion also present reduced accuracy across both \ac{ATE}$_t$ and \ac{ATE}$_r$.
KISS-ICP is the only baseline that competes with RaDiVe on \ac{ATE}$_r$, but due to its inherent reliance on point-to-point correspondence design, its \ac{ATE}$_t$ highly diverges compared with the result from RaDiVe. 
Fig. 5a and Fig. 5b present qualitative results on \texttt{20240113\_1}, presenting others suffer from drift or registration failure, whereas RaDiVe robustly estimates full odometry. 

With the Oculii Eagle, RaDiVe achieves the best \ac{ATE}$_t$ and \ac{ATE}$_r$ in every sequence except {\ac{ATE}$_t$} in \texttt{20240113\_3}, where Doppler-ICP performs better.
However, Doppler-ICP’s instability on other sequences and its high computational cost limit its practical use. 
In contrast, RaDiVe provides robust and stable odometry on all sequences, maintaining real-time performance. 
Fig. 5c and Fig. 5d present qualitative results on \texttt{20240113\_1}, where the other baselines suffer from accumulating orientational drift and fail to revisit the loop. 

When comparing two 4D radars between \tabref{tab:SNAIL_ars_exp} and \tabref{tab:SNAIL_oculii_exp}, 4DRadarSLAM benefits from the denser Oculii Eagle as it was primarily developed for, while RaDiVe outperforms others consistently across both sensors. 
{The comparison also indicates that higher point density does not necessarily yield lower odometry error. The internal point cloud enhancement of Oculii Eagle increases output density but may also affect geometric fidelity, which could be one factor contributing to the observed performance difference between the two sensors.}

{RaDiVe achieves the lowest \ac{RTE} in most comparisons across both sensors. It ranks second to KISS-ICP, with modest differences in translational \ac{RTE} for ARS548 \texttt{20240113\_3} and rotational \ac{RTE} for Oculii Eagle \texttt{20240113\_3}, demonstrating competitive relative error performance.}

\begin{table}[!t]
\centering
\caption{Distance-Bounded NDT in \texttt{Ntu4dradlm}}
\vspace{-2mm}
\label{tab:ablation_ndt}
\resizebox{\columnwidth}{!}{
\begin{tabular}{c|c|c|c|c|c|c}
\toprule
\multicolumn{2}{c|}{ } & \textbf{Ours} & \textbf{Full NDT} & \textbf{GICP} & \textbf{P2P ICP} & \textbf{ICP} \\ \midrule
\multirow{2}{*}{\texttt{loop2}} & ATE$_t$ & \textbf{70.825} & \underline{74.039} & 209.090 & 444.414 & 222.925 \\
 & ATE$_r$ & \textbf{5.928} & \underline{5.995} & 13.145 & 27.275 & 14.420 \\ \midrule
\multirow{2}{*}{\texttt{nyl}} & ATE$_t$ & \textbf{2.970} & {\underline{5.945}} & 8.804 & 7.696 & 7.628 \\ 
 & ATE$_r$ & \textbf{1.472} & \underline{1.961} & 3.192 & 3.021 & 2.961 \\ \bottomrule
\end{tabular}}

\begingroup
    \centering
    \includegraphics[width=0.75\columnwidth]{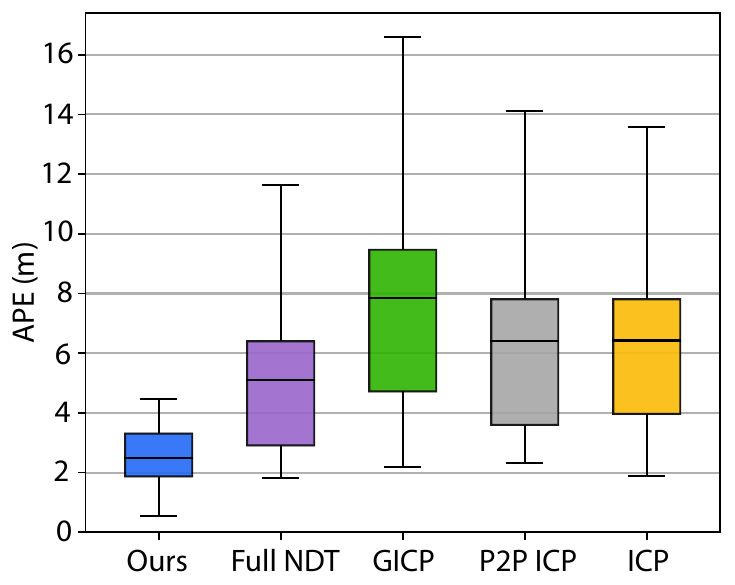}
    \vspace{-4mm}
    \captionof{figure}{An evaluation about different registration methods on \texttt{nyl} sequence in the \texttt{Ntu4dradlm} dataset. The median values of \ac{ATE}$_t$ and standard deviations are (\textbf{2.345}, \textbf{1.102}), (\underline{4.296}, \underline{2.903}), (6.087, 5.255), (4.903, 4.174), and (5.000, 4.061). Distance-bounded NDT presents the lowest median and standard deviation on \ac{ATE}$_t$}
    \vspace{-8mm}
    \label{fig:ndt_iqr}
\endgroup
\end{table}

\subsection{Effect of Distance-Bounded NDT}

We assessed the effectiveness of our distance-bounded \ac{NDT} registration by comparing it to four alternatives on \texttt{Ntu4dradlm} with all other processing held constant: (1) \textbf{Full NDT} (using all possible voxel correspondences without spatial constraints), (2) \textbf{GICP}, (3) \textbf{P2P ICP} (point-to-plane ICP), and (4) \textbf{ICP} (point-to-point ICP), implemented using the Small\_gicp~\cite{small_gicp} library. As shown in \tabref{tab:ablation_ndt}, our approach outperforms all baselines on \texttt{loop2} and \texttt{nyl} for both \ac{ATE}$_t$ and \ac{ATE}$_r$ due to its stable, distance-based correspondence selection. 
Though Full NDT presents a competitive result with our distance-bounded \ac{NDT}, this comes at a cost: Full NDT requires $2\sim3\times$ longer runtime per frame, making it unsuitable for real-time usage. 
These results highlight the balance between accuracy and practical efficiency in RaDiVe.

Interestingly, GICP failed to provide robust odometry with its distribution-based representation. This is because GICP generates a single distribution per point using the \ac{KNN} approach, which results in an overly dense distribution in regions where dense, noisy points exist, leading to poor correspondence. 
Comparably, the \ac{NDT} approach is robust in this situation as a single \ac{NDT} is generated per voxel. 
To evaluate our distance-bounded \ac{NDT} more thoroughly, we drew interquartile range graphs with \ac{APE} among the different registration metrics in \tabref{tab:ablation_ndt} as \figref{fig:ndt_iqr}. 
Both median and standard deviation of \ac{APE} are lowest on our distance-bounded NDT, which supports its robustness and accuracy on a sparse 4D radar point cloud.

\begin{table}[t]
\centering
\caption{{Velocity Uncertainty and Surface Point Extraction}}
\vspace{-2mm}
\label{tab:ablation_pin_unc}
\resizebox{\columnwidth}{!}{
\begin{tabular}{c|c|c|c|c|c|c}
\toprule
\multirow{2}{*}{\textbf{V. Unc.}} & \multirow{2}{*}{\textbf{S. P. E.}} & \multirow{2}{*}{{\textbf{PIN-map}}} & \multicolumn{2}{c|}{\texttt{Ntu. nyl}} & \multicolumn{2}{c}{\texttt{HeRC. P. Lot}} \\
& & & ATE$_t$ & ATE$_r$ & ATE$_t$ & ATE$_r$ \\ \midrule
{\xmark} & {\xmark} & {\xmark} & {97.395} & {65.926} & {3.928} & {6.690} \\
{\cmark} & {\xmark} & {\xmark} & {81.029} & {56.116} & {2.102} & {4.861} \\
\xmark & \xmark & {\cmark} & 78.051 & 74.801 & 30.544 & 68.228 \\ 
\cmark & \xmark & {\cmark} & 19.053 & 12.581 & 3.267 & 5.735 \\
\xmark & \cmark & {\cmark} & \underline{5.872} & \underline{2.649} & \underline{1.206} & \underline{3.052} \\
\cmark & \cmark & {\cmark} & \textbf{2.970} & \textbf{1.472} & \textbf{1.034} & \textbf{1.026} \\
\bottomrule
\end{tabular}}
\vspace{-4mm}
\end{table}

\begin{figure}[!t]
    \centering
    \includegraphics[width=0.8\columnwidth]{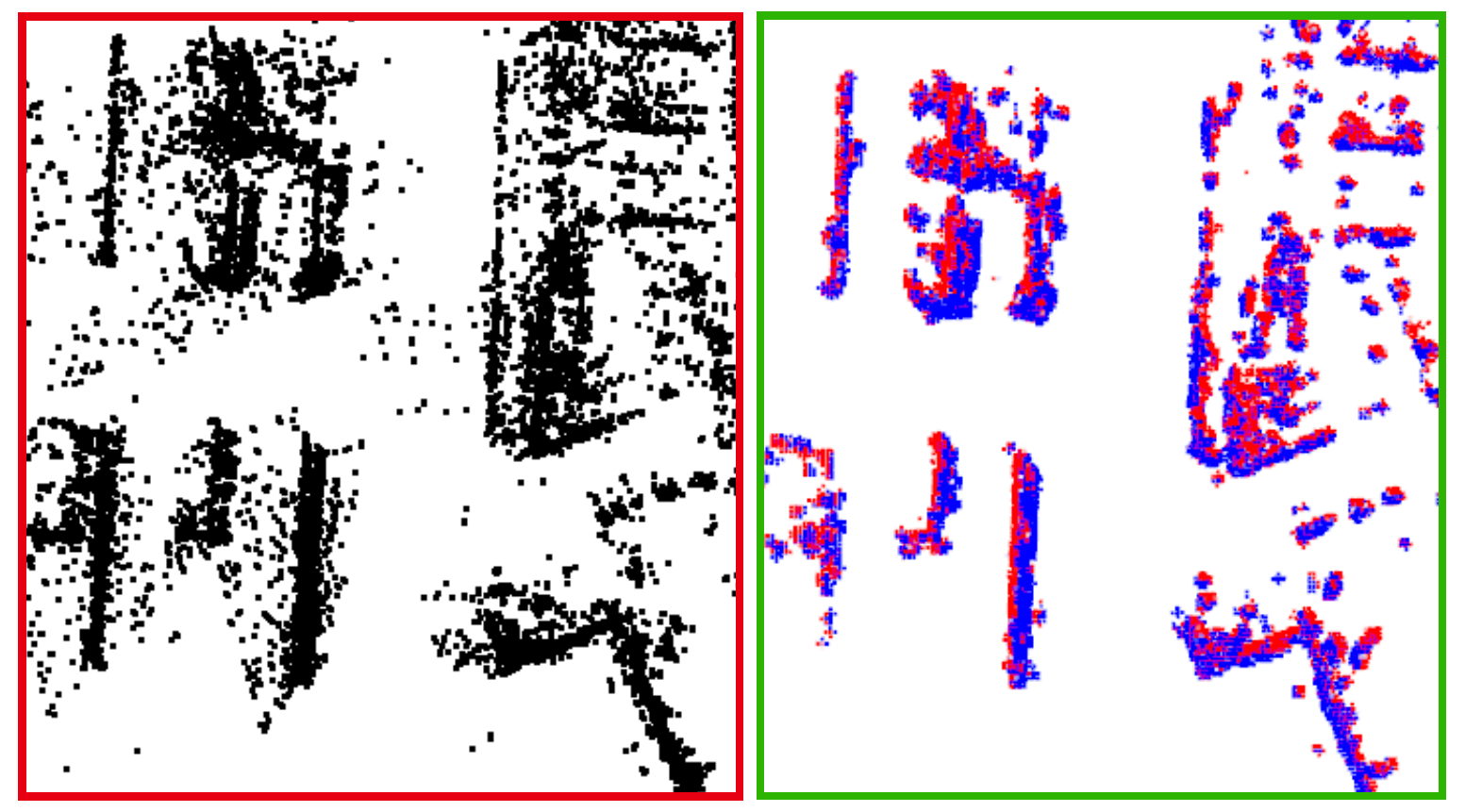}
    \vspace{-4mm}
    \caption{SDF-based Surface Point Extractor on \texttt{Parking Lot}. Noisy points in the left red-bordered figure are deleted in the right green-bordered figure based on the SDF-based surface point extraction. In the right green-bordered figure, the color of the points depicts \ac{SDF} values; blue indicates positive, and red indicates negative \ac{SDF} values.}
    \vspace{-6mm}
    \label{fig:surf_point}
\end{figure}

\subsection{Effect of Uncertainty and Surface Point Extraction}


We evaluate velocity uncertainty (V. Unc.) and surface point extraction (S.P.E.) on \texttt{nyl} and \texttt{Parking Lot Night 3}, and additionally compare PIN-map with simple accumulated global map. Disabling S.P.E. retains PIN-map construction but omits SDF-based filtering. The accumulation baseline uses the same preprocessing, keyframe criteria, local map selection rules, and registration settings. As shown in Table VI, V. Unc. always reduces \ac{ATE} on both sequences. On \texttt{Parking Lot Night 3}, accumulation outperforms PIN-map without S.P.E. This may reflect the influence of off-surface neural points on registration under sparse ARS548. With S.P.E., PIN-map selects points near the SDF zero-level set and achieves lower errors than accumulation. These results support using PIN-map together with S.P.E. for local submap construction.

\subsection{Computational Cost Evaluation}

\begin{figure}[!t]
    \centering
    \includegraphics[width=0.9\columnwidth]{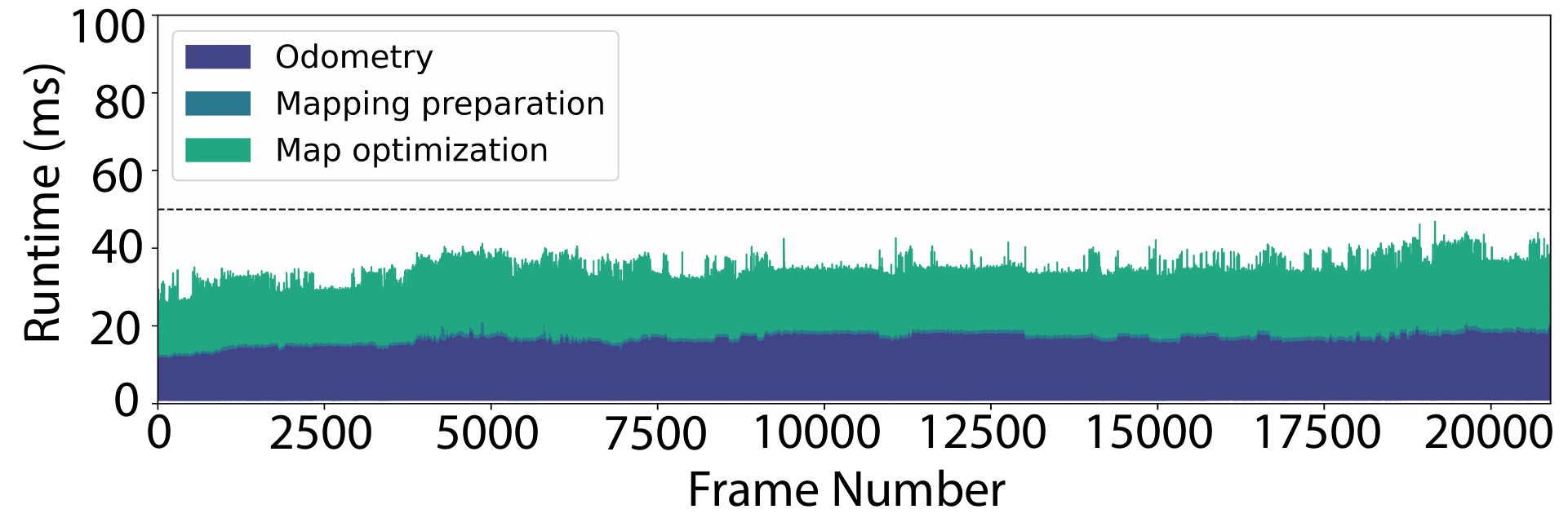}
    \vspace{-4mm}
    \caption{Time consumption on \texttt{Street Day 1}. The horizontal dotted line represents the 50 $\ms$ real-time criterion. }
    \label{fig:time}
    \vspace{-6mm}
\end{figure}

We analyzed the whole runtime of our Python-based pipeline, including \ac{LM} optimization and \ac{NDT} generation. 
Critical functions are compiled using \texttt{@torch.jit.script} decorator for optimized computational efficiency. 
To evaluate the real-time performance of RaDiVe in general, we selected at least one sequence from every public dataset. 
As shown in \tabref{tab:ablation_time}, the average time consumption per frame of RaDiVe is lower than real-time thresholds of $83\ms$ (Oculii Eagle) and $50\ms$ (Continental ARS548) on every chosen sequence. 
Furthermore, a qualitative result on time consumption with \texttt{Street Day 1}, the most extended sequences among \texttt{HeRCULES}, is included in \figref{fig:time}. 
The map optimization time consumption is plotted based on the average time consumption, as the original time consumption graph is jagged due to the keyframe-based mapping. 
All benchmarks were conducted on a system including an Intel i7-11700 CPU, 64GB RAM, and an NVIDIA RTX 3080 GPU, verifying that RaDiVe meets real-time constraints on consumer-grade hardware.

\begin{table}[!t]
\centering
\caption{Average time consumption in multiple public datasets}
\vspace{-2mm}
\label{tab:ablation_time}
\resizebox{0.9\columnwidth}{!}{
\begin{tabular}{c|c||c|c}
\toprule
\textbf{Oculii (83 ms)}  & \ms /frame & \textbf{ARS548 (50 ms)}  & \ms /frame  \\ \midrule
\texttt{loop2} & 72.47 \ms & \texttt{Library} & 49.30 \ms \\
\texttt{nyl} & 74.38 \ms & \texttt{Street} & 42.92 \ms \\ 
\texttt{20240123\_3} & 65.28 \ms & \texttt{20240123\_3} & 49.03 \ms \\ \bottomrule
\end{tabular}}
\vspace{-4mm}
\end{table}

\begin{table}[!b]
\centering
\vspace{-6mm}
\caption{Comparison with LiDAR-based SLAM in \texttt{HeRCULES} with Aeva Aeries II Solid State FMCW LiDAR}
\vspace{-2mm}
\label{tab:ablation_lidar}
\resizebox{\columnwidth}{!}{
\begin{tabular}{cc|cc|cc|cc|cc}
\toprule
& & \multicolumn{2}{c|}{\texttt{Library}} & \multicolumn{2}{c|}{\texttt{Sports Complex}} & \multicolumn{2}{c|}{\texttt{Parking Lot}} & \multicolumn{2}{c}{\texttt{Street}} \\ 
& & ATE$_t$ & ATE$_r$ & ATE$_t$ & ATE$_r$ & ATE$_t$ & ATE$_r$ & ATE$_t$ & ATE$_r$  \\ \midrule
\multirow{3}{*}{\rotatebox{90}{LiDAR}} & \textbf{Fast-LIO2} & 3.982 &\underline{2.088} & 4.635 & 2.514 & \underline{1.982} & \underline{5.382} & 3.139 & 5.023  \\ 
& {\textbf{4DRadarSLAM}} & {26.145} & {14.591} & {22.091} & {8.813} & {22.373} & {53.377} & {23.867} & {15.727} \\
& \textbf{Ours (L)} & \underline{3.425} & \textbf{1.703} & \underline{3.738} & \underline{2.513} & 2.033 & 6.427 & \underline{2.654} & \underline{4.885} \\ \midrule
 & \textbf{Ours (R)} & \textbf{2.179} & 2.901 & \textbf{1.691} & \textbf{1.484} & \textbf{1.034} & \textbf{1.026} & \textbf{2.600} & \textbf{2.571} \\ \bottomrule
\end{tabular}}
\end{table}

\subsection{Cross-Modal Evaluation on FMCW LiDAR}

The importance of a cross-modal framework between 4D radar and \ac{FMCW} LiDAR, which both present 4D point clouds but differ in sensor characteristics, has recently emerged, as presented in Doppler-SLAM~\cite{wang2025doppler}. 
Though RaDiVe is not primarily designed for cross-modal purposes, it can process both types of input data. 
Therefore, we additionally evaluated RaDiVe on dense, high-precision 4D point clouds from the Aeva Aeries II \ac{FMCW} LiDAR in \texttt{HeRCULES}. 
{We compare RaDiVe with FAST-LIO2~\cite{xu2022fast}, a LiDAR-inertial odometry method, and 4DRadarSLAM~\cite{zhang20234dradarslam}, a radar-based SLAM method, using the same FMCW LiDAR data.} {For FMCW LiDAR, we reuse the voxel-based NDT settings employed for NTU Oculii.}

As shown in \tabref{tab:ablation_lidar}, {radar-based RaDiVe achieves lower translational \ac{ATE} on all four evaluated sequences and lower rotational ATE on three of them than both LiDAR-based RaDiVe and FAST-LIO2. On \texttt{Library}, however, both LiDAR-based methods achieve lower rotational \ac{ATE}.} {This comparison characterizes performance under the evaluated configurations, without establishing the best achievable accuracy of each sensor modality.} {LiDAR-based RaDiVe remains competitive with FAST-LIO2 and also achieves lower ATE than 4DRadarSLAM on all sequences, supporting its potential applicability to FMCW LiDAR.}

\section{Conclusion}
\label{sec:conclusion}

In this letter, we presented RaDiVe, a robust 4D radar odometry framework that {improves the accuracy of} radar point cloud registration. Our approach integrates three key components: distance-bounded \ac{NDT} correspondence, velocity-discrepancy point uncertainty, and SDF-based surface point extraction via PIN mapping. These innovations effectively mitigate the challenges of sparse, noisy, and low-precision radar data on odometry estimation. Extensive evaluations on public datasets demonstrate that RaDiVe {outperforms the baselines on most sequences} while maintaining real-time performance. {As future work, we plan to integrate proprioceptive sensors, such as IMU or leg kinematics, to further improve odometry accuracy.}

\section*{Acknowledgment}
The authors used AI-assisted tools to refine \figref{fig:figure_1}a and \figref{fig:figure_1}b. Final verification was performed by the authors. 

\balance
\renewcommand{\bibfont}{\footnotesize}
\bibliographystyle{IEEEtranN} 
\bibliography{string-short,references}

\end{document}